\documentclass[letterpaper, 10 pt, conference]{ieeeconf}  
\makeatletter
\let\NAT@parse\undefined
\makeatother

\IEEEoverridecommandlockouts                              
\usepackage{amssymb}  % assumes amsmath package installed
\usepackage{multirow}

\usepackage{marvosym}
\usepackage{makecell}

\usepackage{graphics} % for pdf, bitmapped graphics files
\usepackage{epsfig} % for postscript graphics files
\usepackage{times} % assumes new font selection scheme installed
\usepackage{stfloats}
\usepackage{enumerate}
\usepackage{booktabs}
\usepackage{todonotes}
\usepackage{algorithm}
\usepackage{algpseudocode}
\usepackage{amsmath}
\usepackage{pifont}
\usepackage{cite}
\usepackage[normalem]{ulem}
\useunder{\uline}{\ul}{}
\usepackage{footnote}
\usepackage{url}

\usepackage[colorlinks=true, linkcolor=blue, citecolor=green, urlcolor=cyan]{hyperref}

\title{\LARGE \bf
Large Language Models Powered Context-aware Motion Prediction \\in Autonomous Driving
}

\author{Xiaoji Zheng, Lixiu Wu, Zhijie Yan, Yuanrong Tang, Hao Zhao\\Chen Zhong\textsuperscript{\Letter}, Bokui Chen\textsuperscript{\Letter}, and Jiangtao Gong\textsuperscript{\Letter}% <-this % stops a space
\thanks{The authors are with the Institute for AI Industry Research, Tsinghua University, Beijing, China. Corresponding Email:
        {\tt\small myheimu@gmail.com, chenbk@tsinghua.edu.cn, gongjiangtao@air.tsinghua.edu.cn}}
}

\begin{document}

\maketitle
% \thispagestyle{empty}
% \pagestyle{empty}

%%%%%%%%%%%%%%%%%%%%%%%%%%%%%%%%%%%%%%%%%%%%%%%%%%%%%%%%%%%%%%%%%%%%%%%%%%%%%%%%
\begin{abstract}

Motion prediction is among the most fundamental tasks in autonomous driving. Traditional methods of motion forecasting primarily encode vector information of maps and historical trajectory data of traffic participants, lacking a comprehensive understanding of overall traffic semantics, which in turn affects the performance of prediction tasks. In this paper, we utilized Large Language Models (LLMs) to enhance the global traffic context understanding for motion prediction tasks. We first conducted systematic prompt engineering, visualizing complex traffic environments and historical trajectory information of traffic participants into image prompts---Transportation Context Map (TC-Map), accompanied by corresponding text prompts. Through this approach, we obtained rich traffic context information from the LLM. By integrating this information into the motion prediction model, we demonstrate that such context can enhance the accuracy of motion predictions. Furthermore, considering the cost associated with LLMs, we propose a cost-effective deployment strategy: enhancing the accuracy of motion prediction tasks at scale with 0.7\% LLM-augmented datasets. Our research offers valuable insights into enhancing the understanding of traffic scenes of LLMs and the motion prediction performance of autonomous driving. The source code is available at \url{https://github.com/AIR-DISCOVER/LLM-Augmented-MTR} and \url{https://aistudio.baidu.com/projectdetail/7809548}.
% We will release our prompt, data, and code for the community soon at: \url{https://github.com/SEU-zxj/LLM-Augmented-MTR}.
%Keywords: large language model, motion prediction, transportation context, prompt engineering
\end{abstract}
\begin{figure}
    \centering
    \includegraphics[width=0.9\linewidth]{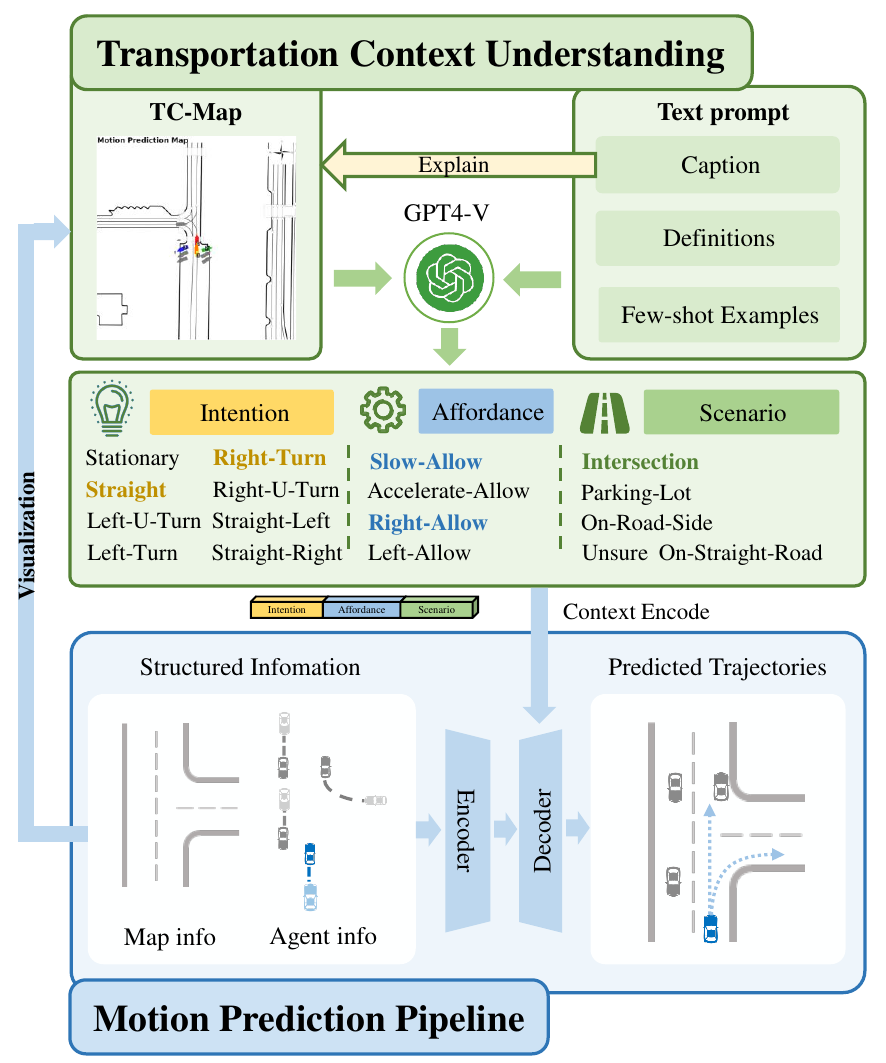}
    \caption{Context-aware Motion Prediction Based on LLMs. We first visualize the structured information of one scenario in the motion prediction dataset, GPT4-V then understand the scenario via a visualized image and well-designed prompt. Finally, GPT4-V outputs transportation context information. This information will be used to augment traditional motion prediction algorithms.}
    \label{fig:main_figure}
\end{figure}

%%%%%%%%%%%%%%%%%%%%%%%%%%%%%%%%%%%%%%%%%%%%%%%%%%%%%%%%%%%%%%%%%%%%%%%%%%%%%%%%
\section{INTRODUCTION}
%Motion prediction 很重要，传统的Motion Prediction使用了什么方法，存在什么问题。

Motion prediction is one of the most important tasks in the field of autonomous driving~\cite{MTR,shi2024mtr++, MultiAgentContextFusion}, which predicts the motion statuses of nearby agents by jointly considering nearby agents and road maps. This information will assist the decision module in making a more robust and safer driving decision.
Therefore, this field already features a plethora of datasets and public competitions. For instance, since 2021, Waymo has been organizing competitions in motion prediction\footnote{\url{https://waymo.com/open/challenges/2021/motion-prediction/}}, attracting models and algorithms that have won championships, such as MTR~\cite{MTR}, MTR++~\cite{shi2024mtr++}, and MGTR~\cite{gan2023mgtr}.
There are three classes of motion prediction methods: goal-based methods~\cite{gu2021densetnt,zhao2021tnt}, direct-regression methods~\cite{ngiam2021scene,varadarajan2022multipath++}, and methods that take the best of both\cite{MTR,shi2024mtr++}. %The goal-based methods adopt goal candidates that cover most motion possibilities, the trade-off of the number of goal candidates is more candidates bring better results while fewer candidates bring less computing resources. The direct-regression methods directly utilize data to regress the final motion trajectories, which can cover all possible cases but coverage slowly. 
Although methods that take the best of both can balance results and computing resources greatly, they still encode information of agents and maps first and then decode to obtain final results. 
The methods above all lack a comprehensive understanding of overall traffic semantics. %In this paper, we tried to augment motion prediction models by providing more direct context semantics of implicit features in the origin context information. How to obtain this context information, the Large Language Model is a good choice.

%LLM很火，不止是语言模型，更是具有common sense的世界模型，有很多将LLM融入自动驾驶的工作
Large Language Model (LLM) is popular in the field of autonomous driving, as not only the Language Model but also the World Model with common sense~\cite{wei2022chain,wang2022self,yao2023tree}. After the release of OpenAI's GPT series of large language models, many works that use GPT to assist driving have been emerging~\cite{GPT-Driver, DriverGPT4, DriveMLM, AgentDriver, jin2023surrealdriver}. These efforts leverage the inferential capabilities of large language models or through experimentation, aiding autonomous driving algorithms in better perception~\cite{DriveLM} and planning~\cite{GPT-Driver, DriveMLM, AgentDriver}, and even facilitating end-to-end decision-making~\cite{wang2023drive, DriverGPT4}.
%然而，让LLM理解有丰富交通交互场景信息是一个有挑战的任务，传统的方法都需要大量的高质量数据来获得驾驶场景的视觉语言模型。例如，XX，xX，XX。这样的成本是非常高昂的。
However, teaching LLM to understand the interactive context among various transportation participants in diverse transportation scenarios is still a challenging task. Although many have applied vision-language models to traffic scenarios, their performance heavily relies on datasets annotated from a first-person perspective~\cite{NuPrompt, NuScenes-QA, Reason2Drive}. For abstract, information-overwhelming, and bird's-eye view (BEV) traffic scenarios, there is a lack of relevant datasets and understanding methodologies. %Lots of high-quality data are needed in traditional methods to obtain the Language Vision Model under driving scenarios \cite{NuPrompt, NuScenes-QA, Reason2Drive, DriveLM}, which is costly.

%\section{Method}

%在本文中，为了让未经fineting的LLM with vision理解交通场景，并且输出运动预测需要的Context信息。我们将交通场景的矢量地图数据和交通参与者历史数据可视化为TC-Map作为图片prompt，并提出了相应text prompt设计。成功的获得LLM生成的交通场景信息后，我们将该信息加入了经典的运动预测算法（引用MTR）中。结果表明，这些Context信息有效地提升了运动预测的准确性。此外，考虑到LLM的成本，我们还提出了一种低成本的部署方案：通过少量的LLM增强数据集，就可规模化地提升运动预测任务的准确性。

In this paper, we propose a new method that enhances the traffic context understanding of motion prediction models using LLMs to make more accurate predictions. To enable an unfine-tuned LLM with vision (e.g. GPT4-V) to understand traffic context and output the necessary context information for motion prediction, we visualize the vector map data and historical data of traffic participants as a Transportation Context Map (TC-Map) to serve as an image prompt and propose a corresponding text prompt design. After successfully obtaining traffic context information generated by the LLM, we integrated this information into a classical motion prediction algorithm---MTR~\cite{MTR}. The results indicate that this context information effectively improves the accuracy of motion prediction. Additionally, considering the cost of LLMs, we also propose a cost-effective deployment strategy: by utilizing 0.7\% LLM-enhanced datasets, we can empower motion prediction performance at scale.

%In this paper, to teach LLM without fine-tuning to understand traffic scenarios and output human-like context information, we first obtain Transportation Context Map (TC-Map) as image prompt by visualizing vector maps and motion history information of agents together and then introduce corresponding text prompt design. After obtaining Transportation Context Information generated by LLM, we integrate it into the traditional motion prediction algorithm \cite{MTR}. The experiment result shows that this context information does improve the accuracy of the motion prediction algorithm. Considering the cost of LVM, we propose a cost-effective deployment strategy, which can empower motion prediction performance at scale by a minimal part of dataset with LLM augmented.
% \vspace{-1.0em}
%所以，本文的贡献可以总结为以下三点：1）XXX；2）XXX；3）XXX。我们的研究对于LLM理解交通场景并增强自动驾驶性能有重要的启示。
Our contributions are as follows: 
% \vspace{-2.5em}
\begin{itemize}
    \item We systematically designed and conducted prompt engineering to enable an unfine-tuned GPT4-V to comprehend complex traffic scenarios involving multiple traffic participants and to output context information such as intention, affordance, and scenario;
    \item We introduced a novel approach that combines the context information outputted by GPT4-V with the classical motion prediction pipeline~\cite{MTR}, and we verified that this method enhances the effectiveness of motion prediction;
    \item We proposed and validated a deployment strategy based on a dataset with a limited amount of LLM-generative context, which reduced the deployment cost of this method.
\end{itemize}
\begin{figure*}
    \centering
    \includegraphics[width=0.85\textwidth]{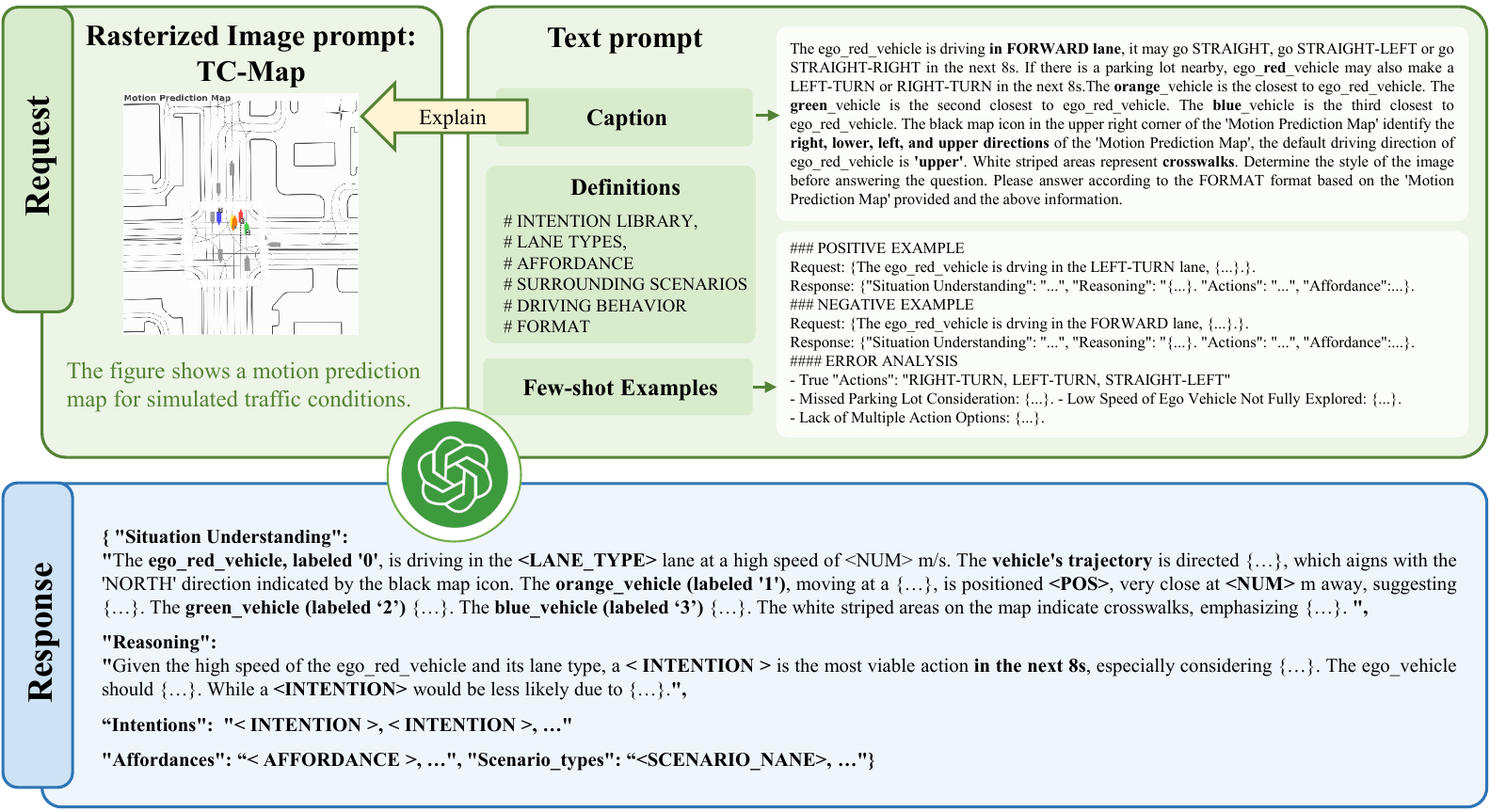}
    \caption{Details of Transportation Context Generation Prompt.}
    \label{fig:prompt}
\end{figure*}
\section{Method}

\begin{figure*}
    \centering
    \includegraphics[width=0.85\textwidth]{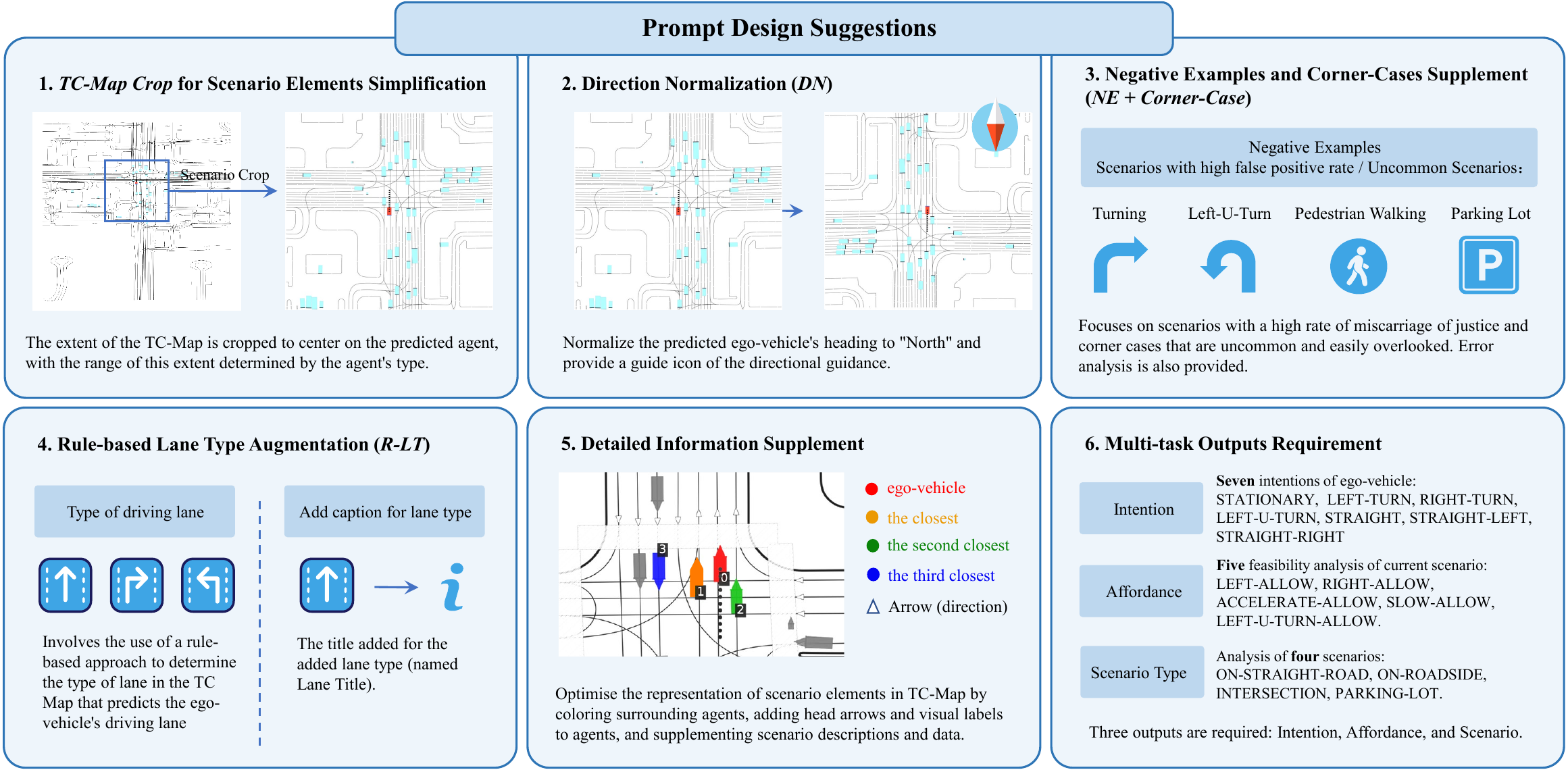}
    \caption{Six Prompt Design Guideline for GPT4-V Understanding Motion Predication Context.}
    \label{fig:promptDesign}
\end{figure*}

% \textbf{1)Request.} As the green part, the input prompt includes TC-Map and Text prompt. Text prompt consists of three parts: Caption provides a textual explanation of the scene information and data within the TC-Map. Definition clarifies the meanings of intention, lane type, affordance, and scenario type, standardizing the analysis process and output format of driving behavior. Few-shot Example offers several examples of both positive and negative.  \textbf{1)Response.} As the blue part, the GPT4-V's answer is divided into understanding, reasoning information and three transportation context.

\subsection{Get Transportation Context Information from GPT4-V}
Translating images directly into motion poses is significantly challenging~\cite{chitta2021neat}. %Inspired by the interpretability of natural language~\cite{TextualExplain4AD, DRAMA} and human inclination for multi-step reasoning, 
We leverage vision-enhanced LLMs to derive transportation context, identifying the agent of interest in the TC-Map as the ego-agent, categorized into vehicles (V), pedestrians (P), and cyclists (C). Illustrated in Fig.~\ref{fig:prompt}, the TC-Map designed based on the Waymo Open Motion Dataset (WOMD) \cite{WOMD} serves as the rasterized image prompt, and together with the text prompt forms the Transportation Context Generation Prompt (TCGP). The caption of the text prompt is obtained by filling in the template with code. Utilizing GPT4-V with this combined input facilitates the extraction of comprehensive context including situation understanding, reasoning, intentions of ego-agent, affordances~\cite{BDD-OIA}, and scenario types. Additionally, six prompt design suggestions have been summarized for input prompts as shown in Fig.~\ref{fig:promptDesign}.

%%%%%%%%%%%%%%%%%%%%polished by GPT%%%%%%%%%%%%%%%%%%%%%%%%%
%\subsection{Extracting Transportation Context from GPT4-V}
%Translating images into motion poses a significant challenge, as noted by Chitta et al.~\cite{chitta2021neat}. We leverage vision-enhanced Large Language Models (LLMs) to derive transportation context, identifying the agent of interest in the Transportation Context Map (TC-Map) as the ego-agent, categorized into vehicles (V), pedestrians (P), and cyclists (C). Illustrated in Fig.~\ref{fig:prompt}, the TC-Map, informed by the Waymo Open Motion Dataset (WOMD) \cite{WOMD}, and a complementary text prompt together constitute the Transportation Context Generation Prompt (TCGP). Utilizing GPT4-V with this combined input facilitates the extraction of comprehensive context including situation understanding, agent intentions, affordances~\cite{BDD-OIA}, and scenario typologies. Furthermore, the prompt design process yielded six strategic recommendations for prompt optimization, detailed in Fig.~\ref{fig:promptDesign}.

% Additionally, six prompt design suggestions have been summarized for input prompts as shown in Fig.~\ref{fig:promptDesign}.

\subsubsection{TC-Map Design Suggestions} Informed by prompt design suggestion 1 and 2. Suggestion 1 minimizes information overload. Drawing inspiration from the MTR Map Collection module~\cite{MTR}, we introduced TC-Map Crop to focus the ego-agent's attention solely on relevant surrounding scenes. Initial GPT4-V tests revealed that dense data, like 11 frames of historical trajectory in JSON format, leading to unclear intentions. Thus, we customized TC-Map dimensions: 120m x 120m for vehicles, 80m x 80m for pedestrians, and 60m x 60m for cyclists, respectively, to filter out extraneous influences on intention discernment. Furthermore, to address the varying orientations of the ego-agent across different scenarios, suggestion 2 proposes normalizing the heading of the ego-agent to ``North'' with a common map guide icon to reduce confusion in direction identification.

%%%%%%%%%%%%%%%%%%%%polished by GPT%%%%%%%%%%%%%%%%%%%
%\subsubsection{TC-Map Design Recommendations} Informed by design recommendations 1 and 2, our first approach minimizes information overload. Drawing inspiration from the MTR Map Collection module~\cite{MTR}, we employed a TC-Map Cropping technique, focusing the ego-agent's attention solely on relevant surrounding scenes. Initial GPT4-V tests revealed that dense data, like 11 frames of historical trajectory in JSON format, overwhelms the model, leading to unclear intentions. Thus, we customized TC-Map dimensions: 120m x 120m for vehicles and 80m x 80m and 60m x 60m for pedestrians and cyclists, respectively, to filter out extraneous influences on intention discernment. Furthermore, to address the varying orientations of the ego-agent across different scenarios, our second recommendation standardizes the ego-agent's heading to "North," employing a universal map guide icon to streamline direction recognition.

\subsubsection{Text Prompt Design Suggestions}  Addressing design suggestions 3 and 4, we introduce a structured approach to example creation. Suggestion 3 involves categorizing examples into positive and negative instances. Positive examples ensure a diverse and random generation of intentions. Negative examples are bad cases selected from GPT4-V's responses during the quantitative experiment of prompts, focusing on scenarios with high misjudgment rates and uncommon, easily overlooked corner cases. Additional error analysis is provided and placed closer to the end of the negative example. Suggestion 4 consists of two sentences, the first sentence contains the ego-agent's driving lane type in TC-Map determined by a rule-based algorithm, and the second sentence provides explanations for the added lane type. They aids GPT4-V in ruling out implausible intentions by assuming compliance with traffic regulations.

%%%%%%%%%%%%%%%%%%%%%%%%%%%%%%%%polished by GPT%%%%%%%%%%%%%%%%%%%%%%%%%%%
%\subsubsection{Text Prompt Design Enhancements} Addressing design suggestions 3 and 4, we introduce a structured approach to example creation. Suggestion 3 involves categorizing examples into positive and negative instances. Positive examples ensure a diverse and random generation of intentions, while negative examples—derived from GPT4-V's erroneous outputs during trials—highlight scenarios prone to misinterpretation or overlook rare, complex cases. We conclude the prompt with an error analysis segment. Suggestion 4 involves detailing the ego-agent's lane type on the TC-Map, determined via a rule-based method, followed by a rationale for the chosen lane. This dual-sentence structure aids GPT4-V in ruling out implausible intentions by assuming compliance with traffic regulations.

\subsubsection{Detailed Information Supplement} Building on prompt design suggestion 5, we visualize more scenario elements using WOMD's comprehensive scenario details, including road edges and crosswalks. For agents, we optimize the visualization of different scenario elements in the TC-Map by adding indicative head arrows and visual labels. Moreover, this suggestion includes integrating dynamic elements such as the ego-agent's speed and the relative speeds, positions, and distances of nearby agents, enriching the context for more accurate motion prediction.

%%%%%%%%%%%%%%%%%%%%%%%%polished by GPT%%%%%%%%%%%%%%%%%%%%%%%%%%%%
%\subsubsection{Detailed Information Integration} Building on suggestion 5, we harness comprehensive scenario details from WOMD, including road edges and crosswalks, to refine TC-Map visualizations. This entails enhancing scenario elements with directional arrows and agent labels for clearer interpretation. Moreover, this strategy includes integrating dynamic elements such as the ego-agent's speed and the relative speeds, positions, and distances of nearby agents, enriching the context for more accurate motion prediction.

\subsubsection{Multi-task Outputs Requirement} Matching to prompt design suggestion 6, we delineated seven intention categories for the ego-agent inspired by the official evaluation tool of Waymo \cite{WOMD}. \textit{STRAIGHT-LEFT} and \textit{STRAIGHT-RIGHT} anticipate potential lane changes, indicating the complexity of intentions. Predicting these intentions often requires GPT4-V to utilize rasterized images, the ego-agent's position, heading, and speed, the state of surrounding vehicles, and the direction of nearby lane lines predicting these two intentions is highly challenging. Additionally, our approach includes the creation of five affordance types and four scenario types based on an understanding of comprehensive scenario analyses. These three types of information collectively provide rich, human-like transportation context information to enhance the accuracy of the motion prediction task.

%%%%%%%%%%%%%%%%%%%%%polished by GPT%%%%%%%%%%%%%%%%%%%%%%%%
%Aligning with the sixth prompt design suggestion, we delineated seven intention categories for the ego-agent, a methodology inspired by Waymo's official assessment tool. Notably, the intentions \textit{STRAIGHT-LEFT} and \textit{STRAIGHT-RIGHT} anticipate potential lane changes, underscoring the intricate nature of intention prediction. This complexity necessitates that GPT4-V interprets a variety of inputs including rasterized images, as well as the ego-agent’s spatial orientation, velocity, and the dynamics of surrounding traffic and lane directions. Furthermore, our approach includes the creation of five affordance types and four scenario classifications, all informed by comprehensive scene analyses. Collectively, these diverse data types furnish a nuanced, human-like understanding of transportation contexts, significantly bolstering motion prediction accuracy.

%Except for intentions, we design another 2 types of context information (see Fig.~/ref{fig:main_figure}). The affordance~\cite{BDD-OIA} provides XXX information, and the scenario provides YYYY information. These three types of information collectively providing a rich, human-like transportation context information to enhance t

\subsection{Augment Motion Prediction Models via Transportation Context Information}
\subsubsection{Motion Transformer}
We selected the Motion Transformer (MTR)~\cite{MTR}, the state-of-the-art model in the WOMD Challenge 2022, as our foundational model. Leveraging the Transformer architecture~\cite{Transformer}, MTR encodes comprehensive scenario data—including agent movements and map features—before decoding this information to predict motion trajectories. This process underpins our model's ability to accurately anticipate future positions.

%%%%%%%%%%%%%%%%polished by GPT%%%%%%%%%%%%%%%%%%
%We selected the Motion Transformer (MTR)~\cite{MTR}, the state-of-the-art model in the WOMD Challenge 2022, as our foundational model. Leveraging the Transformer architecture~\cite{Transformer}, MTR encodes comprehensive scenario data—including agent movements and map features—before decoding this information to predict motion trajectories. This process underpins our model's ability to accurately anticipate future positions.

\subsubsection{Integrate with Transportation Context Information}
To enrich our motion prediction model with transportation context, we categorized context information into three types: intention, affordance, and scenario, each represented as natural language descriptors (e.g. [\textit{Straight}, \textit{Right-Turn}], [\textit{Slow-Allow}, \textit{Right-Allow}], and [\textit{Intersection}] in Fig.~\ref{fig:main_figure}).
%As the number of each type of context information is fixed, we use one-hot vectors to express the context information. In addition, because the description words in the list are not exclusive, we allow several `1's to appear in one one-hot vector.
These descriptors are encoded as one-hot vectors, allowing for multiple active elements to reflect the non-exclusive nature of these descriptors. Notably, affordances and scenarios are directly encoded, while intentions receive weighted encoding to prioritize them.

More specifically, for affordance and scenario information generated by LLM, we simply encode the description words into one-hot vectors:
\begin{align}
    A &= OneHotEncode(AffordanceList), \label{eq:affordance} \\
    S &= OneHotEncode(ScenarioList), \label{eq:scenario}
\end{align}
where $A \in \mathbb{R}^8$ and $S \in \mathbb{R}^4$ (see Fig.~\ref{fig:main_figure}). 
%The number of ground truth intentions is only one, but the number of description words in the intention information list is not strictly one.
LLM gives all possible intentions and the order of different description words stands for different possibilities. So, when we encode intention information into a one-hot vector, we set different weights for different intentions in the list:
\begin{equation}
    I = WeightedOneHotEncode(IntentionList). \label{eq:intention}
\end{equation}
The weight for i-th word in the list is:
\begin{equation}
    weight_i = length(IntentionInfoList) - i, \label{eq:weight}
\end{equation}
where $I \in \mathbb{R}^5$ (see Fig.~\ref{fig:main_figure}), and each intention's weight inversely correlates with its position in the list, emphasizing more probable intentions. So that the first word in the list has the maximal weight, and the last word holds the minimal weight.

%%%%%%%%%%%%%%%%%%%%polished by GPT%%%%%%%%%%%%%%%%%%%%%%%%%
%To enrich our motion prediction model with transportation context, we categorized context information into three types: intention, affordance, and scenario, each represented as natural language descriptors (e.g., [\textit{Straight}, \textit{Right-Turn}] for intention). These descriptors are encoded as one-hot vectors, allowing for multiple active elements to reflect the non-exclusive nature of these descriptors. Notably, affordances and scenarios are directly encoded ($A$ and $S$, respectively), while intentions receive weighted encoding to prioritize them based on likelihood:

% XXXXX

%where each intention's weight inversely correlates with its position in the list, emphasizing more probable intentions.

%To integrate this context with our Motion Transformer (MTR) model, we first amalgamate these encoded vectors and align them with the dimensional requirements of the model's query vectors. Utilizing cross-attention mechanisms, the enriched context (TC) informs the decoding process, improving prediction accuracy by leveraging nuanced, LLM-generated insights into transportation dynamics."

To improve the performance of motion prediction and reduce the cost of computing resources, cluster-based intention points were provided in the process of decoding as prior knowledge \cite{MTR}. Inspired by this, we also choose to integrate the transportation context information generated by LLM in the process of decoding.

The query content $Q \in \mathbb{R}^{K \times D}$ includes all scenario information in the data, where $K$ stands for the number of trajectories that need to be predicted and $D$ is the dimension of the feature. We first concatenate different types of context information and align the dimensions of traffic context information with query content, then integrate the context information generated by LLM via cross-attention after query content has been initialized.
\begin{align}
    TC &= MLP([I, A, S]), \label{eq:align1} \\
    TC &= RepeatFirstDim(TC, K), \label{eq:align2} \\
    Q_{tc} &= CrossAttn(q = Q_0, k = TC, v = TC), \label{eq:crossattn}
\end{align}
where $TC \in \mathbb{R}^D$ in Eq. \ref{eq:align1} after multi-layer perceptron, and $TC \in \mathbb{R}^{K \times D}$ after repeating $K$ times in Eq. \ref{eq:align2}. When conducting cross-attention, we use initialized query content $Q_0$ as query, and $TC$ for both key and value in Eq. \ref{eq:crossattn}.

The following pipeline keeps all the same with MTR.

\begin{table*}[]
 \centering
\caption{\textbf{Effects of Different Components in TCGP}}
\label{tab:effect_componts}
\resizebox{\linewidth}{!}{
\begin{tabular}{cccccc|c|c|c}
\toprule
\textbf{\textit{Sugg 1}} & {\color[HTML]{000000} \textbf{\textit{Sugg 2}}} & \textbf{\textit{Sugg 3}} & \textbf{\textit{Sugg 4}} & \multicolumn{2}{c|}{\textbf{\textit{Sugg 5}}} &  &  &  \\ 
\cmidrule(lr){1-1} \cmidrule(lr){2-2} \cmidrule(lr){3-3} \cmidrule(lr){4-4} \cmidrule(lr){5-6}
\textbf{TC-Map Crop} & \textbf{DN} & \textbf{NE+Corner-Case} & \textbf{R-LT} & \textbf{TCGP-Opt} & \textbf{SPD-Add} & \multirow{-2}{*}{\textbf{ACC(\textit{1st}-\(I\))}} & \multirow{-2}{*}{\textbf{ACC}} & \multirow{-2}{*}{\textbf{ACC(\textit{S,S-L,S-R})}} \\
\midrule
\textcolor{red}{\ding{55}}& \textcolor{red}{\ding{55}} & \textcolor{red}{\ding{55}} & \textcolor{red}{\ding{55}} & \textcolor{red}{\ding{55}} & \textcolor{red}{\ding{55}} & 0.3551 & 0.5140 & 0.5794 \\
\textcolor{green}{\ding{52}} & \textcolor{red}{\ding{55}} & \textcolor{red}{\ding{55}} & \textcolor{red}{\ding{55}} & \textcolor{red}{\ding{55}} & \textcolor{red}{\ding{55}} & 0.4579 & 0.5981 & 0.7383 \\
\textcolor{green}{\ding{52}} & \textcolor{green}{\ding{52}} & \textcolor{red}{\ding{55}} & \textcolor{red}{\ding{55}} & \textcolor{red}{\ding{55}} & \textcolor{red}{\ding{55}} & 0.4205 & 0.6635 & 0.7757 \\
\textcolor{green}{\ding{52}} & \textcolor{green}{\ding{52}} & \textcolor{green}{\ding{52}} & \textcolor{red}{\ding{55}} & \textcolor{red}{\ding{55}} & \textcolor{red}{\ding{55}} & 0.4112 & 0.6822 & 0.8037 \\
\textcolor{green}{\ding{52}} & \textcolor{green}{\ding{52}} & \textcolor{green}{\ding{52}} & \textcolor{green}{\ding{52}} & \textcolor{red}{\ding{55}} & \textcolor{red}{\ding{55}} & 0.5607 & 0.8317 & 0.9252 \\
\textcolor{green}{\ding{52}} & \textcolor{green}{\ding{52}} & \textcolor{green}{\ding{52}} & \textcolor{green}{\ding{52}} & \textcolor{green}{\ding{52}} & \textcolor{red}{\ding{55}} & 0.5514 & \textbf{0.8691} & 0.9252 \\
\textcolor{green}{\ding{52}} & \textcolor{green}{\ding{52}} & \textcolor{green}{\ding{52}} & \textcolor{green}{\ding{52}} & \textcolor{green}{\ding{52}} & \textcolor{green}{\ding{52}} & \textbf{0.5794} & \textbf{0.8691} & \textbf{0.9345} \\ 
\bottomrule
\end{tabular}
}
\vspace{2pt}
\footnotesize{\\
\textbf{\textit{Sugg}}: Prompt design suggestion. \textbf{TCGP-Opt}: Coloring agents surrounding the ego-agent in different colors to distinguish them easily, optimizing the visualization of scenario elements in the TC-Map and supplementing the corresponding scenario description in the text prompt. \textbf{SPD-Add}: Adding information about the speed of the ego-agent, the speeds of surrounding agents, and their relative positions and distances to the ego-agent in the text prompt. \textbf{ACC}: The accuracy of intentions output by GPT4-V. When GPT4-V outputs multiple intentions, if at least one intention matches the ground truth without additional clarification, it is considered correct. \textbf{(\textit{1st}-\(I\))}: Taking only the first intention from the list of intentions output by GPT4-V. \textbf{(\textit{S,S-L,S-R})} means consolidating the three intentions of \textit{STRAIGHT}, \textit{STRAIGHT-LEFT}, and \textit{STRAIGHT-RIGHT} into a single \textit{STRAIGHT} category.
}
\end{table*}

%%%%%%%%%%%%%%%%%%polished by GPT%%%%%%%%%%%%%%%%%%%%
%Sugg (Prompt Design Suggestion): General strategies proposed for designing prompts to guide GPT4-V in generating transportation context.

%TCGP-Opt (Transportation Context Generation Prompt Optimization): Involves color-coding agents around the ego-agent for easier differentiation, enhancing TC-Map scenario visualizations, and elaborating scenario descriptions within the text prompt for improved clarity.

%SPD-Add (Speed and Position Details Addition): Enriches the text prompt with details regarding the ego-agent's speed, the velocities of surrounding agents, and their relative positions and distances from the ego-agent, aiming to provide a comprehensive situational context.

%ACC (Accuracy): Measures the precision of intentions predicted by GPT4-V. Accuracy is affirmed when at least one of the multiple intentions output by GPT4-V aligns with the actual (ground truth) scenario, without necessitating further specification.

%1st−I (First Intention Consideration): Considers only the foremost intention in GPT4-V's list of generated intentions for assessing prediction accuracy.

%S,S−L,S−R (Intention Consolidation): Simplifies the analysis by amalgamating the intentions 'STRAIGHT', 'STRAIGHT-LEFT', and 'STRAIGHT-RIGHT' into a singular 'STRAIGHT' category for streamlined evaluation."

\section{Experiement}
\subsection{TCGP Generated Transportation Context Information}

\begin{figure}
    \centering
    \includegraphics[width=6cm]{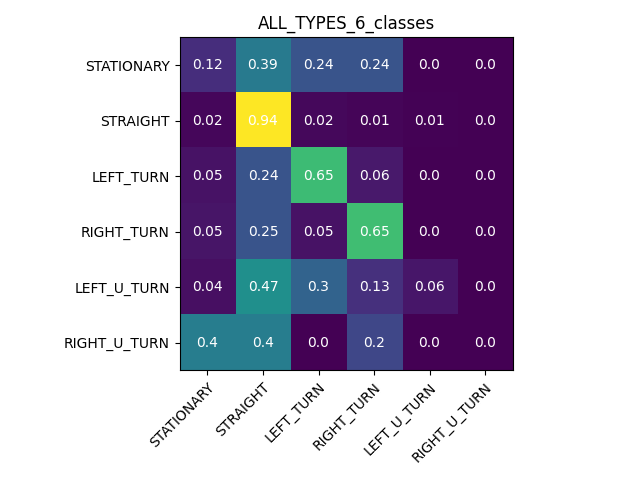}
    \caption{Confusion Matrix of Intention Generated from GPT4-V}
    \label{fig:ConsusionMatrix6Class}
\end{figure}

\begin{table}[]
% \centering
\caption{\textbf{Data and Intention Accuracy of Three Ego-Agent Types} }
\label{tab:data_and_intention_acc}
\resizebox{\linewidth}{!}{
\begin{tabular}{c|c|ccc}
\toprule
method & type & ACC(\textit{1st}-\(I\)) & ACC & ACC(\textit{S,S-L,S-R}) \\
\midrule
\multirow{4}{*}{MTR}
 & V & 0.8097 & 0.9442 & 0.9635 \\
 & P & 0.6906 & 0.9495 & 0.9564 \\
 & C & 0.7226 & 0.8822 & 0.9321 \\
 & AVG & 0.7644 & 0.9321 & 0.9532 \\
\midrule
\multirow{4}{*}{GPT4-V} 
 & V & 0.6224  & 0.8482 & 0.9080 \\
 & P & 0.6222  & 0.8038 & 0.8133 \\
 & C & 0.5838 & 0.7857 & 0.9030 \\
 & AVG & 0.6145 & 0.8259 & 0.8865 \\
\bottomrule
\end{tabular}
}

\end{table}

% \begin{table}[]
% \caption{\textbf{Data and Intention Accuracy of Three Ego-Agent Types.} }
% \label{tab:data_and_intention_acc}
% \resizebox{\linewidth}{!}{
% \begin{tabular}{c|ccc}
% \toprule
%  & ACC(\textit{1st}-\(I\)) & ACC & ACC(\textit{S,S-L,S-R}) \\
% \midrule
% Vehicle & 0.6224  & 0.8482 & 0.9080 \\
% Pedestrian & 0.6222  & 0.8038 & 0.8133 \\
% Cyclist & 0.5838 & 0.7857 & 0.9030 \\
% All & 0.6145 & 0.8259 & 0.8865 \\
% \bottomrule
% \end{tabular}
% }
% \end{table}

% The data for the transportation context information generation were randomly sourced from the WOMD validation set, comprising about 15,000 TC-Maps samples based on V : P : C = 9 : 3 : 3.
\textbf{Data distribution.} The data set used to generate transportation context information was randomly selected from the WOMD validation set, consisting of approximately 15,000 TC-Map samples distributed according to a vehicle (V) to pedestrian (P) to cyclist (C) ratio of 9:3:3. We create a confusion matrix for the results of intention in Tab.~\ref{tab:data_and_intention_acc} as shown in Fig.~\ref{fig:ConsusionMatrix6Class}. For simplification, the categories \textit{STRAIGHT-LEFT} and \textit{STRAIGHT-RIGHT} were amalgamated into a single \textit{STRAIGHT} category. The matrix's rows denote the actual intentions, while its columns represent the intentions as predicted by GPT4-V. Observing the main diagonal of the confusion matrix, it's not hard to see that \textit{STRAIGHT} has the highest prediction accuracy, followed by \textit{LEFT-TURN} and \textit{RIGHT-TURN} predictions. 
The comparatively lower accuracy observed for \textit{STATIONARY} intentions stems from its correct identification only at a zero speed of the ego-agent, despite being labeled as \textit{STATIONARY} at minimal speeds. 
Ego-agents intended for \textit{LEFT-U-TURN} are frequently classified as \textit{LEFT-TURN}, a consequence of the WOMD map's left-turn lanes occasionally accommodating straight movements, thus increasing the likelihood of \textit{LEFT-U-TURN} and \textit{LEFT-TURN} being predicted as \textit{STRAIGHT}. Similarly, \textit{RIGHT-U-TURN} ego-agents are often mistaken as \textit{STRAIGHT}, and unexpectedly, they are frequently classified as \textit{STATIONARY}. This anomaly primarily arises because intentions like \textit{RIGHT-U-TURN} are most common among pedestrians, whose speed and walking direction can change abruptly, reflecting the challenge of precisely capturing the dynamic intentions of pedestrians.
Despite GPT4-V's intention prediction accuracy being lower than that of the MTR model, the integration of GPT4-V's output still contributes valuable context for motion prediction, enhancing overall prediction accuracy, as evidenced in Tab.~\ref{tab:PerformanceCompare}.

\textbf{Ablation Study.} We perform ablation studies on TCGP to understand the effectiveness of each component of our prompt in Tab.~\ref{tab:effect_componts}. 
The data comprising 107 TC-Maps samples for the ablation experiment were randomly sourced from the validation set of WOMD. We have the following observations: \textbf{1)} TCGP achieves optimal performance when fully equipped with all its components, with every module playing a role in predicting intentions. \textbf{2)} TC-Map Crop enables GPT4-V to allocate more attention to the areas that should be focused on when predicting the future intentions of the ego-agent, significantly enhancing the accuracy of intention prediction. \textbf{3)} R-LT is crucial because it provides the type of lane in which the ego-agent is located, and the type of lane often determines the future direction of the ego-agent. These observations underscore the multifaceted contributions of TCGP's components to the overall effectiveness of intention prediction, highlighting the importance of a holistic approach in optimizing predictive performance.

\subsection{LLM Augmented Motion Prediction}
% \subsubsection{generate similar transportation context information}

% reason: 1) training needs data (specifically, XX tokens are needed for each agent. The price of OpenAI GPT4-V is XXX\$/XX tokens. X\$ for each agent. XXX data in Waymo Open Motion Dataset (WOMD). Roughly, we need \$ to generate all LLM hints).

\textbf{Dataset and metrics.} We evaluated our LLM-augmented motion prediction methods using the extensive WOMD dataset. Task requirements include predicting 6 future motion trajectories over 8 seconds, based on 1 second of historical data. The official WOMD evaluation tool calculates key metrics, notably the mean Average Precision (mAP), which serve as crucial benchmarks in the official leaderboard .

%%%%%%%%%%%%%%%%%%%%polished by GPT%%%%%%%%%%%%%%%%%%%%%%
%We evaluated our LLM-augmented motion prediction methods using the extensive WOMD dataset. Task requirements include predicting 6 future motion trajectories over 8 seconds, based on 1 second of historical data. The official WOMD evaluation tool calculates key metrics, notably the mean Average Precision (mAP) and Miss Rate (MR), which serve as crucial benchmarks in the official leaderboard

\textbf{Data generation.} To manage the high cost of generating transportation context information for WOMD's 2 million agents (approximately \$0.1 per agent, totaling nearly \$200,000), we implemented a cost-effective strategy. Initially, LLM-generated transportation context for a small subset of WOMD was expanded to the remainder via a nearest neighbor algorithm, utilizing Euclidean distance for similarity assessment (see Alg.~\ref{alg:knn}). This approach efficiently scales LLM augmentation while maintaining data integrity.

%%%%%%%%%%%%%%%%%%polished by GPT%%%%%%%%%%%%%%%%
%To manage the high cost of generating transportation context information for WOMD's 2 million agents (approximately $0.1 per agent, totaling nearly $200,000), we implemented a cost-effective strategy. Initially, LLM-generated transportation context for a small subset of WOMD was expanded to the remainder via a nearest neighbor algorithm, utilizing Euclidean distance for similarity assessment. This approach efficiently scales LLM augmentation while maintaining data integrity

\renewcommand{\algorithmicrequire}{\textbf{Input:}}  % Use Input in the format of Algorithm
\renewcommand{\algorithmicensure}{\textbf{Output:}} % Use Output in the format of Algorithm

\begin{algorithm}[h]
  \caption{Generate Scaled LLM Augmented Training Data with Minimal LLM Augmented Data}
  \label{alg:knn}
  \begin{algorithmic}[1]
    \Require
      $T$: the overall dataset without LLM augmented;
      $Encoder$: scenario info encoder;
      $LLM$: large language model.
    \Ensure
      $T^{*}$: the overall dataset with LLM augmented.
    \State initial $T_1, T_2 \leftarrow split(T)$, where $\left\vert T_1 \right\vert \gg \left\vert T_2 \right\vert$
    \State $TC_1 \leftarrow []$, where $TC_1$ is Transportation Context info
    \State $TC_2 \leftarrow LLM(T_2)$ 
    \State $F_1 \leftarrow Encoder(T_1)$, where $F_1$ is feature vector list
    \State $F_2 \leftarrow Encoder(T_2)$
    \For{agent i in $T_1$}
        \State $f_i \leftarrow F_1[i]$, where $f_i$ is a feature vector 
        \State $j \leftarrow \textbf{NearestNeighbor}(f_i, F_2)$
        \State $TC_1[i] = TC_2[j]$
    \EndFor
    \State $T^{*}_1 \leftarrow Concatenate(T_1, TC_1)$
    \State $T^{*}_2 \leftarrow Concatenate(T_2, TC_2)$
    \State $T^{*} \leftarrow Concatenate(T^{*}_1, T^{*}_2)$
  \end{algorithmic}
\end{algorithm}

% \subsubsection{performance of different methods}
% analyze: 1) hints of LLM do improve the performance of the motion prediction model 2) in the process of inference, real LLM hints information can also be integrated into the model, which brings more accurate results as the performance shows.
% \usepackage{multirow}
\textbf{Performance comparison.} We compared our LLM augmented motion prediction model with MTR on the validation set of WOMD. We trained both models on the 20\% WOMD dataset. Our main results are in Table \ref{tab:PerformanceCompare}. 
% V, P, C are abbreviations of vehicle, pedestrian, cyclist, respectively. 
Our LLM-augmented model exhibited superior performance over the MTR model across all agent types (vehicles, pedestrians, cyclists) on the WOMD validation set. Notably, the average mAP saw an enhancement of 0.95\%, underscoring the augmented model's effectiveness.

%%%%%%%%%%%%%%%%%%%polished by GPT%%%%%%%%%%%%%%%%%%%%%%
%Our LLM-augmented model exhibited superior performance over the MTR model across all agent types (vehicles, pedestrians, cyclists) on the WOMD validation set. Notably, the average mAP saw an enhancement of 0.95%, underscoring the augmented model's effectiveness.

\begin{table}[h]
\caption{\textbf{Performance on the validation set of WOMD} }
\label{tab:PerformanceCompare}
\centering{
    \begin{tabular}{c|c|cccc}
    \toprule
    method                & type       & mAP$\uparrow$   & minADE$\downarrow$ & minFDE$\downarrow$ & MR$\downarrow$ \\
    \midrule
    \multirow{4}{*}{MTR}  & V    & 0.3862          & 0.8257 & 1.6789 & 0.1809   \\
                          & P & 0.3587          & 0.3825 & 0.8088 & 0.0935   \\
                          & C    & 0.2848          & 0.7985 & 1.6380 & 0.2212   \\
                          & AVG        & 0.3432          & 0.6689 & 1.3752 & 0.1652   \\
    \midrule
    \multirow{4}{*}{+LLM} & V    & \textbf{0.3954} & 0.8147 & 1.6205 & 0.1751   \\
                          & P & \textbf{0.3754} & 0.3830 & 0.8070 & 0.0934   \\
                          & C    & \textbf{0.2924} & 0.8102 & 1.6464 & 0.2306   \\
                          & AVG       & \textbf{0.3527} & 0.6693 & 1.3580 & 0.1664   \\
    \bottomrule
    \end{tabular}
}
\vspace{2pt}
\footnotesize{\\
The best scores are expressed in \textbf{bold}.
}
\end{table}
% \vspace{-1.0em}

% \subsubsection{ablation experiment of LLM augmented motion prediction}
% \textbf{Ablation experiment.} We study the effectiveness of each component of the transportation context generated by LLM. To efficiently conduct the ablation experiment, we randomly sampled 5\% scenarios (about 24k scenarios) from the WOMD training set. Both models are evaluated on the validation set of WOMD, for the LLM augmented model, similar transportation context information is used. Table \ref{tab:MTR+LLMAblationExperiment} shows our experiment results. The mAP of models with a single type of transportation context information augmented (see lines 2,3,4) verifies that each type of transportation context does empower the accuracy of motion prediction. More specifically, the intention-augmented model obtains the best mAP score while the scenario-augmented model obtains the best minADE, minFDE, and MR scores. Suppose all types of transportation context information are provided for the motion prediction model. In that case, the same mAP score with the intention-augmented model will be obtained, and 2ed best minADE and minFDE will also obtained. The mAP score of the final model is increased by 1.37\% compared with MTR.

\textbf{Ablation Study.} This study evaluates the contribution of each component within the transportation context generated by the LLM toward enhancing motion prediction accuracy. An ablation experiment was conducted by randomly selecting 5\% of scenarios (approximately 24,000 scenarios) from the WOMD training set to train a model. The evaluation involved comparing the baseline model against the LLM-augmented model on the minimal part of the WOMD validation set with traffic context information generated by LLM. The results are presented in Table \ref{tab:MTR+LLMAblationExperiment}. I, A, and S are abbreviations of intention, affordance, and scenario, respectively. V-mAP, P-mAP, and C-mAP mean mAP scores for vehicle, pedestrian, and cyclist, respectively. The results highlight the incremental value added by incorporating various types of transportation context information.  Incorporation of all context types resulted in the highest mAP scores, marking a 1.67\% increase over the baseline MTR model, thereby validating the efficacy of LLM-augmented context in enhancing motion prediction.
%%%%%%%%%%%%%%%%%%%%%%%%%%%polished by GPT%%%%%%%%%%%%%%%%%%%%%%%%
%The ablation study, focused on a 5% sample of the WOMD training set, illuminated the individual and collective contributions of intention, affordance, and scenario information towards motion prediction accuracy. Incorporation of all context types resulted in the highest mAP scores, marking a 1.67% increase over the baseline MTR model, thereby validating the efficacy of LLM-augmented context in enhancing motion prediction.

\begin{table}[h]
\caption{\textbf{Ablation Experiment of \\
LLM Augmented Motion Prediction} }
\label{tab:MTR+LLMAblationExperiment}
\centering
\begin{tabular}{c|ccc|cccc}
\hline
method & I         & A          & S         & mAP$\uparrow$             & V-mAP$\uparrow$           & P-mAP$\uparrow$           & C-mAP$\uparrow$           \\ \hline
MTR    & -         & -          & -         & 0.2970          & 0.3300          & 0.3507          & 0.2102          \\
-      & \checkmark & -          & -         & 0.3066          & 0.3377          & 0.3400          & 0.2422          \\
-      & -         & \checkmark  & -         & 0.2995          & 0.3386          & 0.3277          & 0.2322          \\
-      & -         & -          & \checkmark & 0.3059          & 0.3341          & 0.3564          & 0.2272          \\
-      & \checkmark & \checkmark  & -         & 0.3030          & 0.3294          & 0.3458          & 0.2337          \\
-      & \checkmark & \textbf{-} & \checkmark & 0.3134          & 0.3318          & \textbf{0.3692} & 0.2392          \\
-      & -         & \checkmark  & \checkmark & 0.3117          & \textbf{0.3389} & 0.3554          & 0.2409          \\
+LLM   & \checkmark & \checkmark  & \checkmark & \textbf{0.3137} & 0.3385          & 0.3549          & \textbf{0.2476} \\
\bottomrule
\end{tabular}
\vspace{2pt}
\footnotesize{\\
The highest mAP score for each type of agent and the average mAP score is expressed in \textbf{bold}.
}
\end{table}
\vspace{-1.0em}
\section{Conclusions}
In this paper, through systematic prompt engineering, we utilized the common sense and reasoning abilities of LLMs to extract human-like global context information from complex traffic scene motion predictions. By integrating these contexts into the traditional motion prediction pipeline, we enhanced the accuracy of trajectory prediction. We also proposed a cost-effective deployment strategy and verified its effectiveness. 

Through meticulous prompt design, we achieved an impressive accuracy rate, underscoring the remarkable capacity of LLMs to comprehend complex and detailed transportation scenarios. To our knowledge, our work is pioneering in incorporating BEV-like TC-Maps into LLM prompts, offering a novel perspective for LLMs to interpret driving scenarios. Our experimental findings suggest that integrating transportation context information significantly enhances motion prediction capabilities. %This method demonstrates robust scalability, potentially augmenting various motion prediction models. %Notably, the frequent invocation of LLM APIs incurs substantial costs and delays, posing challenges for practical deployment. Future work will explore the adoption of open-source LLMs to mitigate expenses and reduce data transfer times associated with network communications.
It is noteworthy that enabling LLMs to understand complex scene information remains challenging. Significant opportunities exist for exploring the integration of rich contextual information with motion prediction models. Our preliminary results suggest considerable potential for future development in this research area.

\bibliographystyle{IEEEtran}
\bibliography{IEEEfull,root}
% \bibliographystyle{IEEEtran}
% \bibliography{root}
% \bibliography{root}

\end{document}